\pdfoutput=1

\documentclass[11pt]{article}

\usepackage[]{ACL2023}

\usepackage{times}
\usepackage{latexsym}

\usepackage[T1]{fontenc}

\usepackage[utf8]{inputenc}

\usepackage{microtype}

\usepackage{inconsolata}
\usepackage{xurl}

\usepackage{graphicx}
\usepackage{floatrow}
\usepackage{subcaption}
\usepackage{multirow}
\usepackage{amssymb}
\usepackage[flushleft]{threeparttable}
\usepackage{amsmath}
\usepackage{relsize}
\usepackage{booktabs}

\title{Recurrent Attention Networks for Long-text Modeling}

\author{Xianming Li\textsuperscript{\rm 1}$^\dagger$,
    Zongxi Li\textsuperscript{\rm 2}$^\dagger$\thanks{\ \ Corresponding author; $^\dagger$ Equal contribution.}, Xiaotian Luo\textsuperscript{\rm 1}, Haoran Xie\textsuperscript{\rm 3}, Xing Lee\textsuperscript{\rm 1}, \\ \textbf{Yingbin Zhao\textsuperscript{\rm 1}, Fu Lee Wang\textsuperscript{\rm 2}, Qing Li\textsuperscript{\rm 4}} \\
  \textsuperscript{\rm 1} Ant Group, Shanghai, China \\
  \textsuperscript{\rm 2} School of Science and Technology, Hong Kong Metropolitan University, Hong Kong SAR \\
  \textsuperscript{\rm 3} Department of Computing and Decision Sciences, Lingnan University, Hong Kong SAR \\
  \textsuperscript{\rm 4} Department of Computing, Hong Kong Polytechnic University, Hong Kong SAR \\
  \texttt{\{niming.lxm,lxt267638,lx250976,zyb166123\}@antgroup.com}\\
  \texttt{\{zoli, pwang\}@hkmu.edu.hk}, 
  \texttt{hrxie@ln.edu.hk}, 
  \texttt{qing-prof.li@polyu.edu.hk}\\
 }

\begin{document}
\maketitle

\begin{abstract}

Self-attention-based models have achieved remarkable progress in short-text mining. However, the quadratic computational complexities restrict their application in long text processing. Prior works have adopted the chunking strategy to divide long documents into chunks and stack a self-attention backbone with the recurrent structure to extract semantic representation. Such an approach disables parallelization of the attention mechanism, significantly increasing the training cost and raising hardware requirements. Revisiting the self-attention mechanism and the recurrent structure, this paper proposes a novel long-document encoding model, Recurrent Attention Network (RAN), to enable the recurrent operation of self-attention. Combining the advantages from both sides, the well-designed RAN is capable of extracting global semantics in both token-level and document-level representations, making it inherently compatible with both sequential and classification tasks, respectively. Furthermore, RAN is computationally scalable as it supports parallelization on long document processing. Extensive experiments demonstrate the long-text encoding ability of the proposed RAN model on both classification and sequential tasks, showing its potential for a wide range of applications.

\end{abstract}

\section{Introduction}
Recently, self-attention-based neural networks, such as Transformer \cite{NIPS2017_transformer}, GPT \cite{radford2018gpt1,radford2019gpt2,brown2020gpt3}, and BERT family \cite{DevlinCLT19BERT, roberta-liu-2019, albert-lan-2020}, have demonstrated superior text encoding ability in many natural language processing (NLP) tasks with the help of large-scale pretraining. These models have set state-of-the-art benchmarks in classification tasks like text categorization \cite{li2021merging} and sentiment analysis \cite{naseem2020transformer,zongxi_kbs_2021,li2023novel}, and sequential tasks like question answering \cite{DBLP:conf/acl/LeeCT19, DBLP:conf/emnlp/KarpukhinOMLWEC20} and information extraction \cite{DBLP:conf/emnlp/LiLDYLH21,DBLP:conf/acl/0001ZL22}. The time and space complexities of self-attention computation are $O(n^2)$ with respect to the sequence length, making it computationally expensive to encode long texts. 
Therefore, BERT models adopt an absolute positional encoding strategy to manage computational overhead. However, such a setting makes the BERT models unable to handle texts longer than 512 tokens, restricting their application in realistic scenarios like processing user comments, news articles, scientific reports, and legal documents with arbitrary lengths. 

Current works focus on two solutions to enable self-attention-based models for handling longer texts. The first solution reduces the computing complexity of self-attention from quadratic to linear by approximating its softmax operation \cite{longformer-iz-2020,DBLP:conf/iclr/ChoromanskiLDSG21,hua2022transformer}. These models can handle relatively long texts within the hardware capacity but also suffer from a performance drop \cite{schlag2021linear,blockformer-delesley-2022}. Another solution is to divide the long document into chunks shorter than 512 tokens so that pretrained BERT models can be applied \cite{tobert-raghavendra-2019,blockformer-delesley-2022}. However, as the chunks are individually encoded, the resulted representations do not contain the crucial contextual information for sequential tasks. While a special recurrent mechanism can handle sequential tasks \cite{blockformer-delesley-2022}, it cannot produce a document-level representation for classification, limiting their generality as none works for both classification and sequential tasks. Additionally, introducing recurrent modules disables the parallel computing feature, leading to unscalable implementation.

\begin{figure}
    \centering
    \subfloat[RNN]{\includegraphics[]{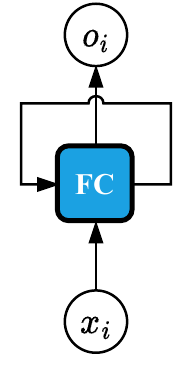}}
    \subfloat[RAN]{\includegraphics[]{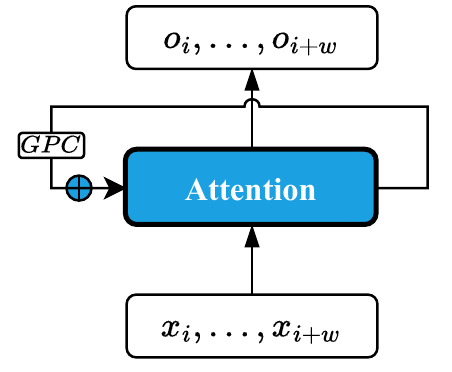}}
    \caption{The architectures of (a) RNN and (b) RAN. The basic encoding unit of RNN is the fully-connected (FC) layer, and that of RAN is the self-attention layer. RNN is a token-level recurrent architecture, while the proposed RAN is a window-level recurrent model.}
    \label{simple-network-figure}
\end{figure}

To address the aforementioned issues, this paper proposes the \textbf{R}ecurrent \textbf{A}ttention \textbf{N}etwork (RAN)\footnote{The code is available at \url{https://github.com/4AI/RAN}.}, a novel model architecture supporting recurrent self-attention operation over long sequences, enabling global dependency extraction and long-term memory. RAN iterates through the sequence by non-overlapping windows. Unlike token-level recurrent architectures such as LSTM \cite{hochreiter1997long-lstm} and GRU \cite{chung2014empirical-gru}, RAN applies positional multi-head self-attention (pMHSA) on a window area to extract local dependency. To propagate the information forward, the RAN model extracts the global perception cell (GPC) vector from the self-attention representation of the current window. The GPC vector is then concatenated with tokens in the next window as the input of the self-attention layer. The new GPC vector will be passed to the subsequent windows with residual connection to alleviate the gradient vanishing \cite{residual-he-2016} and updated in the same manner. 
Figure \ref{simple-network-figure} depicts the difference between the recurrent neural network (RNN) and our proposed RAN.

The function of the GPC vector is twofold. First, like the [CLS] token in BERT, the GPC vector is a window-level contextual representation. But unlike the [CLS] token, the GPC vector is only applied to the self-attention layer, and no special token is inserted during text preprocessing. 
Second, the GPC vector, resembling the state cell in a recurrent architecture, maintains a long-distance memory over the sequence. For each window, the attended GPC vector encodes an aggregated representation of all the previous windows, which enables the window-level self-attention to perceive global semantics. With the help of a well-designed memory review mechanism, the GPC vector from the last window can be used as a document-level representation and serve the classification tasks. Meanwhile, the memory review mechanism enhances the token representations of RAN in the sequence, encoding both contextual and global information, which can be leveraged for sequential tasks such as language modeling (LM) and named entity recognition (NER).


We pretrain the RAN model using a masked language modeling (MLM) objective from scratch, which outperforms other pretrained baselines in long document classification.
The RAN framework also supports auto-regressive LM and achieves the lowest perplexity score compared with state-of-the-art language models on the WikiText-103 dataset. Furthermore, we apply RAN to different downstream tasks via finetuning and observe consistent improvements compared to baseline models.

RAN solely relies on self-attention, and no LSTM-style gate is involved when propagating information via GPC vectors. Therefore, RAN is computationally efficient as it supports parallelized GPU computing. Although the memory complexity is still quadratic, it is regarding the window size $W$ rather than the whole text length $L$, where $W \ll L$. Nevertheless, the window size can be adjusted based on hardware availability to achieve a relatively larger batch size for better training.

In summary, our contribution is to devise the RAN model for long document processing. RAN allows for parallelization on GPU and provides the interfaces for serving both classification and sequential tasks. With pretraining, RAN can outperform the BERT-based models in various tasks.

\section{Related Work}
This section reviews the relevant works focusing on sequence modeling in NLP, especially long document processing. RNNs are widely used for sequential modeling by recursively updating a state cell to maintain a long-distance memory. Traditional recurrent networks, such as LSTM \cite{hochreiter1997long-lstm} and GRU \cite{chung2014empirical-gru}, use the fully-connected layer as the basic encoding unit and apply the gate mechanism to update state memory. The recurrent operation is conducted on the token level, which is inefficient as such a framework cannot compute parallelly on GPU. Besides, it might suffer from gradient vanishing for long sequences during the backpropagation phase \cite{blockformer-delesley-2022}.

Self-attention models are powerful in global representation learning. However, applying self-attention in long document processing is intractable due to the quadratic time and memory complexities. To address this issue, some works \cite{longformer-iz-2020,DBLP:conf/iclr/ChoromanskiLDSG21,hua2022transformer} attempt to reduce the computing complexity of self-attention from quadratic to approximately linear complexity. \citet{longformer-iz-2020} propose a drop-in replacement of the softmax operation in self-attention with a sparse attention mechanism. Similarly, \citet{DBLP:conf/iclr/ChoromanskiLDSG21} rely on prior knowledge like sparsity and low-rankness to efficiently estimate the full-rank attention. However, these approaches face a trade-off between efficiency and accuracy, as approximations may lead to a performance drop \cite{schlag2021linear,blockformer-delesley-2022}. 

Other works leverage the power of full-rank self-attention as backbones, such as pretrained BERT and RoBERTa. These works cope with the token-length limitation with different strategies. \citet{cogltx-ding-2020} propose CogLTX framework to generate a brief summary of the document. The short summary is used for the classification task employing BERT. However, it is inevitable to lose information in the length compression. \citet{tobert-raghavendra-2019} segment the long text into smaller chunks so that BERT can be then used. A recurrent layer is employed to obtain the document-level representation upon chunk-level representations. These models can be applied for the classification task but cannot handle sequential tasks because of losing crucial contextual and sequential information. \citet{blockformer-delesley-2022} adopt the chunking strategy and devise a specifically-designed gate mechanism to obtain token-level representations for sequential tasks. Similarly, \citet{didolkar2022temporal} propose a Transformer-based temporal latent bottleneck for image classification, reinforcement learning, and text classification, in which temporal states are updated using a recurrent function across chunks. In each Transformer block, temporal states update the chunk-level representation by cross-attention layers interleaved with self-attention layers. In general, these models with BERT backbones cannot simultaneously handle classification and sequential tasks for long documents. Meanwhile, the RNN-style gate architecture does not support parallel computing, so the computing efficiency is also impaired. 

Our proposed RAN achieves recurrent operation of the self-attention model and hence supports parallelization. Moreover, like the traditional RNN architecture, RAN can produce both token-level and document-level representations, which can be leveraged for both sequential and classification tasks.

\section{Methodology}

\begin{figure}[ht]
    \centering
    \includegraphics[width=0.99\textwidth]{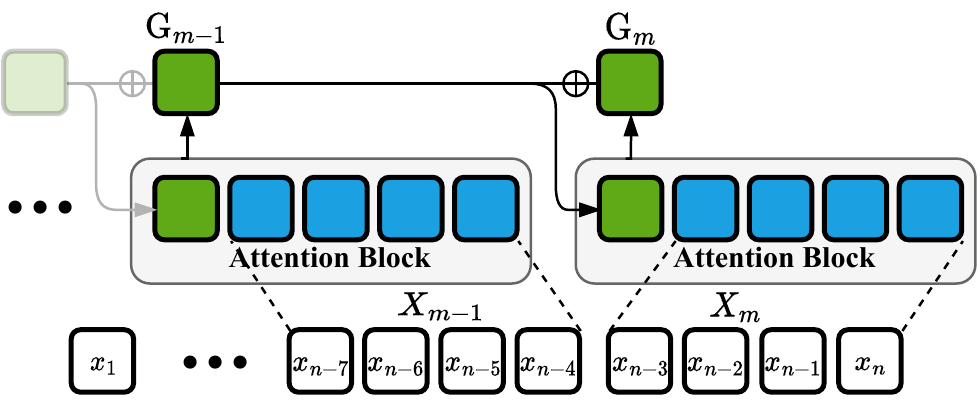}
    \caption{The generic framework of RAN.}
    \label{ran-model-figure}
\end{figure}

\begin{figure*}[ht]
    \centering
    \includegraphics[width=0.87\textwidth]{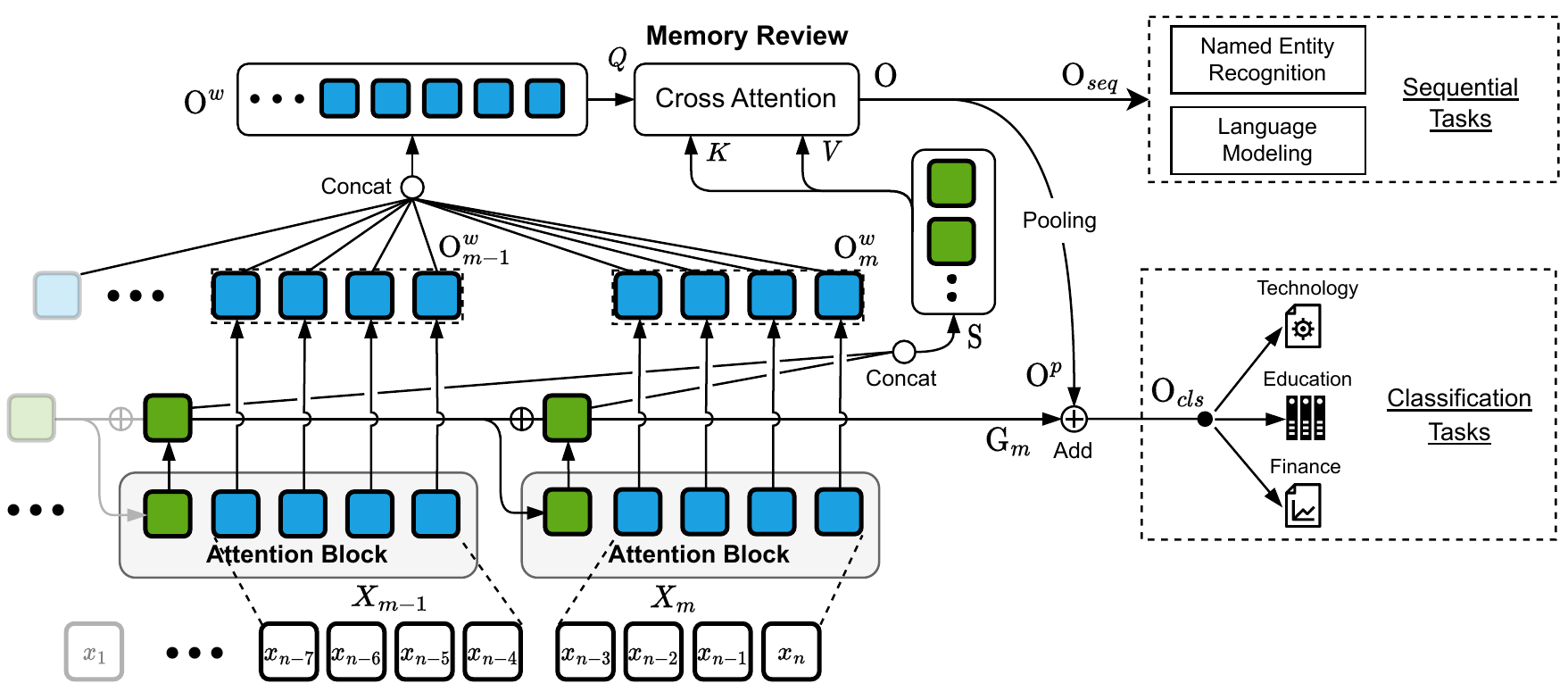}
    \caption{The frameworks of RAN with memory review on different tasks.}
    \label{ran-model-for-tasks-figure}
\end{figure*}


This section introduces the proposed RAN framework in terms of its components. Figure \ref{ran-model-figure} depicts the structure of the basic RAN module. In RAN, the primary encoder is the pMHSA, encoding the GPC vector and the current input with the rotary positional information carried \cite{DBLP:journals/corr/abs-2104-09864}. The GPC vector is employed to propagate information through the sequence.


\subsection{Input Layer}

We first employ the padding operation for the input documents to keep a uniform length $L$. Then we map each word into a $D$-dimensional continuous space and obtain the word embedding $\mathbf{x}_i \in \mathbb{R}^D$. The word vectors are concatenated to form the model input: $\mathbf{X} = [\mathbf{x}_1, \mathbf{x}_2, \dots, \mathbf{x}_L] \in \mathbb{R}^{L \times D}$. To feed the text into the RAN, we chunk the input document into $m=\mathrm{ceil}(\frac{L}{W})$ windows, where $W$ is the window size. We use $\mathbf{X}_{i} \in \mathbb{R}^{W \times D}$ to denote the $i$-th window input. In RAN, the GPC vector $\mathcal{G}_0 \in \mathbb{R}^D$ is initialized to $\mathbf{0}$ by default, following the common operation in RNN. We parameterize the GPC vector with the layer normalization \cite{layernorm-lei-2016} as follows:

\begin{equation}
    \mathbf{G}_{0} = \mathrm{LayerNorm}(\mathbf{W}_{g} \mathcal{G}_0) \in \mathbb{R}^D.
\end{equation}

\subsection{Positional Multi-Head Self-Attention}

As the positional space of long documents is prohibitively large, it is not feasible to use absolute positional embedding following Transformer families \cite{NIPS2017_transformer} in long document processing. Hence, we follow \citet{DBLP:journals/corr/abs-2104-09864,DBLP:journals/corr/abs-2204-02311,DBLP:journals/corr/abs-2204-06745} to incorporate the local positional information of the current window and leverage rotary position information as follows:

\begin{equation}
    \mathrm{pMHSA}(\mathbf{X}_i) = \mathbf{W} [\mathrm{Att}_1 (\mathbf{X}_i);...;\mathrm{Att}_h (\mathbf{X}_i)] + b, 
\end{equation}
and
\begin{equation}
    \begin{split}
        \mathrm{Att}_j (\mathbf{X}_i) &= \mathrm{SoftMax}(\frac{\mathrm{RP}(\mathbf{Q}_j) \cdot \mathrm{RP}(\mathbf{K}^{\mathrm{T}}_j)} {\sqrt{d^k}} \\ 
         & \quad\quad\quad\quad +\mathbf{M}) \mathbf{V}_j \\
        \mathbf{Q}_j &= \mathbf{W}^q_j \mathbf{X}_i + b^q_j \\
        \mathbf{K}_j &= \mathbf{W}^k_j \mathbf{X}_i + b^k_j \\
        \mathbf{V}_j &= \mathbf{W}^v_j \mathbf{X}_i + b^v_j,
    \end{split}
\end{equation}
where $\mathrm{Att}_j(\cdot)$ is the $j$-th head of pMHSA, $h$ denotes the head size, $[;]$ means the concatenation operation, $\mathbf{M}$ is the attention mask to adapt to different sequential tasks, and $\mathrm{RP}(\cdot)$ stands for the rotary position function \cite{DBLP:journals/corr/abs-2104-09864}.

\subsection{Encoding and Updating Layer}
To encode the $i$-th window, we concatenate the GPC vector from the previous window $\mathbf{G}_{i-1}$ and the current window input $\mathbf{X}_{i} \in \mathbb{R}^{W \times D}$ to form the model input $\mathbf{X}^{in}_{i} =[\mathbf{G}_{i-1}; \mathbf{X}_{i}] \in \mathbb{R}^{(1 + W) \times D}$. A layer normalization layer is applied to normalize the input.

We then apply the pMHSA to encode the concatenated input to obtain the outputs of the current window:
\begin{equation}
    \mathbf{O}_{i} = \mathrm{pMHSA}(\mathbf{X}^{in}_{i}).
\end{equation}

After encoding, we extract the updated GPC vector $\mathbf{G}'_{i}$ and the output corresponding to the tokens in the window:
\begin{equation}
    \begin{split}
        \mathbf{G}'_i &= \mathrm{SN}(\mathbf{O}_{i}^{[1:2]}) \in \mathbb{R}^{D} \\
        \mathbf{O}^w_{i} &= \mathrm{SN}(\mathbf{O}_{i}^{[2:1+W]}) \in \mathbb{R}^{W \times D},
    \end{split}
    \label{updated-eq}
\end{equation}
where $^{[start:end]}$ is the tensor slice operation, and $\mathrm{SN}(X) = \frac{X - X_{mean}}{\sigma}$ stands for the standard normalization.
To alleviate the gradient vanishing issue in modeling long sequences, we employ residual connection to connect the current GPC vector with the previous one, then pass it to a layer normalization layer to normalize the updated GPC vector,
\begin{equation}
    \mathbf{G}_{i} = \mathrm{LayerNorm}(\mathbf{G}'_i + \mathbf{G}_{i-1}).
\end{equation}
The updated GPC vector $\mathbf{G}_i \in \mathbb{R}^D$ will be propagated to the next window.


\subsection{Memory Review and Output Layer}
After encoding all windows, we can obtain the sequence output by concatenating all window outputs, as follows:
\begin{equation}
    \mathbf{O}^w = [\mathbf{O}^w_{1};\mathbf{O}^w_{2};...;\mathbf{O}^w_{m}] \in \mathbb{R}^{L \times D},
\end{equation}
where $m$ is the number of windows. $\mathbf{O}^w$ has the same shape as the input $\mathbf{X}$. To prevent history forgetting in handling long sequences, this paper proposes a novel memory review mechanism. Specifically, we first concatenate all updated GPC vectors to produce the history states vector:

\begin{equation}
    \mathbf{S} = [\mathbf{G}_1;\mathbf{G}_2;...;\mathbf{G}_m] \in \mathbb{R}^{m \times D}.
\end{equation}
We compute the cross attention of the concatenated output and the historical memory states to obtain the final output:

\begin{equation}
    \begin{split}
        \mathbf{O} &= \mathrm{SoftMax}(\frac{\mathbf{Q} \mathbf{K}^T}{\sqrt{d^k}}) \mathbf{V} \\
        \mathbf{Q} &= \mathbf{W}^{q} \mathbf{O}^w + b^{q} \\
        \mathbf{K} &= \mathbf{W}^{k} \mathbf{S} + b^{k} \\
        \mathbf{V} &= \mathbf{W}^{v} \mathbf{S} + b^{v}.
    \end{split}
\end{equation}
This procedure mimics the human behavior of reviewing key points after reading an article, the way that humans naturally consolidate information and reinforce memory.

The sequence output $\mathbf{O} \equiv \mathbf{O}_{seq} \in \mathbb{R}^{L \times D}$ can be used for sequential tasks like NER. Although the GPC vector of the last window, $\mathbf{G}_m$, can serve as the document representation, it may lose crucial semantics and long-term memory during the propagation. Therefore, we also add the memory review mechanism to RAN for classification tasks by generating $\mathbf{O}_{clf}$:  
\begin{equation}
    \mathbf{O}_{clf} = \mathbf{W}^g \mathbf{G}_m + \mathbf{W}^o \mathbf{O}^{p} + b^o,
\end{equation}
where $\mathbf{O}^{p}$ is the pooling of the output $\mathbf{O}$ over time sequence. Our empirical results show that the max pooling works better than the average pooling in classification tasks. Therefore, we adopt max pooling to obtain $\mathbf{O}^{p}$:
\begin{equation}
    \mathbf{O}^p = \mathrm{MaxPooling}(\mathbf{O}).
    \label{eq:mxpool}
\end{equation}

Figure \ref{ran-model-for-tasks-figure} provides a visual illustration of the implementations for both classification and sequential tasks. It is noticeable that the model parameters of RAN are shared across all windows, allowing for efficient computation and reduced memory usage. Particularly, RAN supports multiple sequential tasks with different attention masks. For instance, it employs a causal attention mask \cite{NIPS2017_transformer} for LM tasks and a prefix causal attention mask \cite{unilm-li-2019} for the seq2seq tasks to prevent forward information exposure.


\section{Experiment}

\subsection{Datasets and Evaluation Metrics}

To comprehensively evaluate the model performance, we conduct experiments on three major tasks: text classification (TC), NER, and LM.

\begin{figure*}[ht]
    \centering
    \subfloat[Text length distribution boxplot. 
    ]{\includegraphics[width=0.515\textwidth]{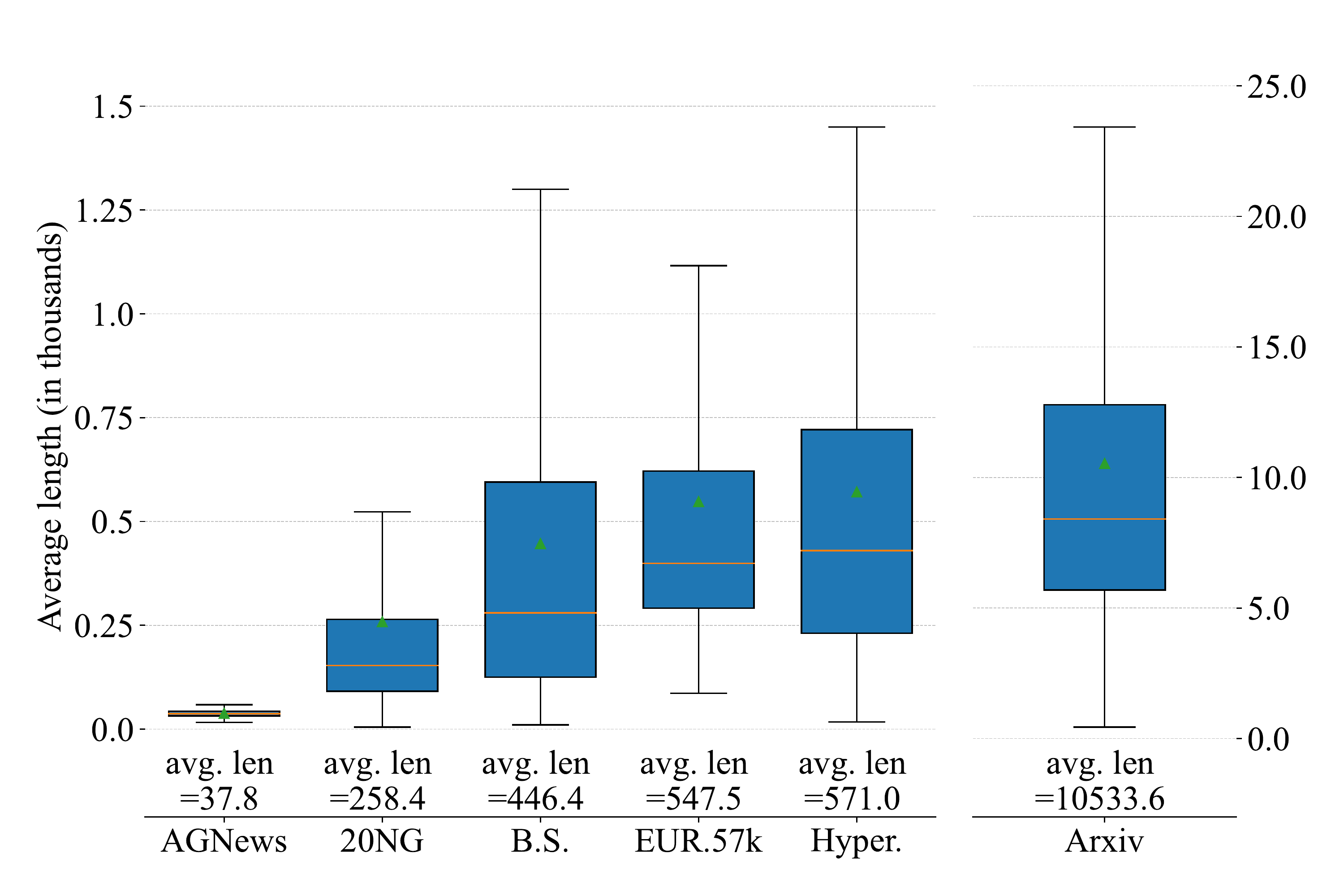}}
    \subfloat[Ratio of the long texts (length $>512$).]{\includegraphics[width=0.42\textwidth]{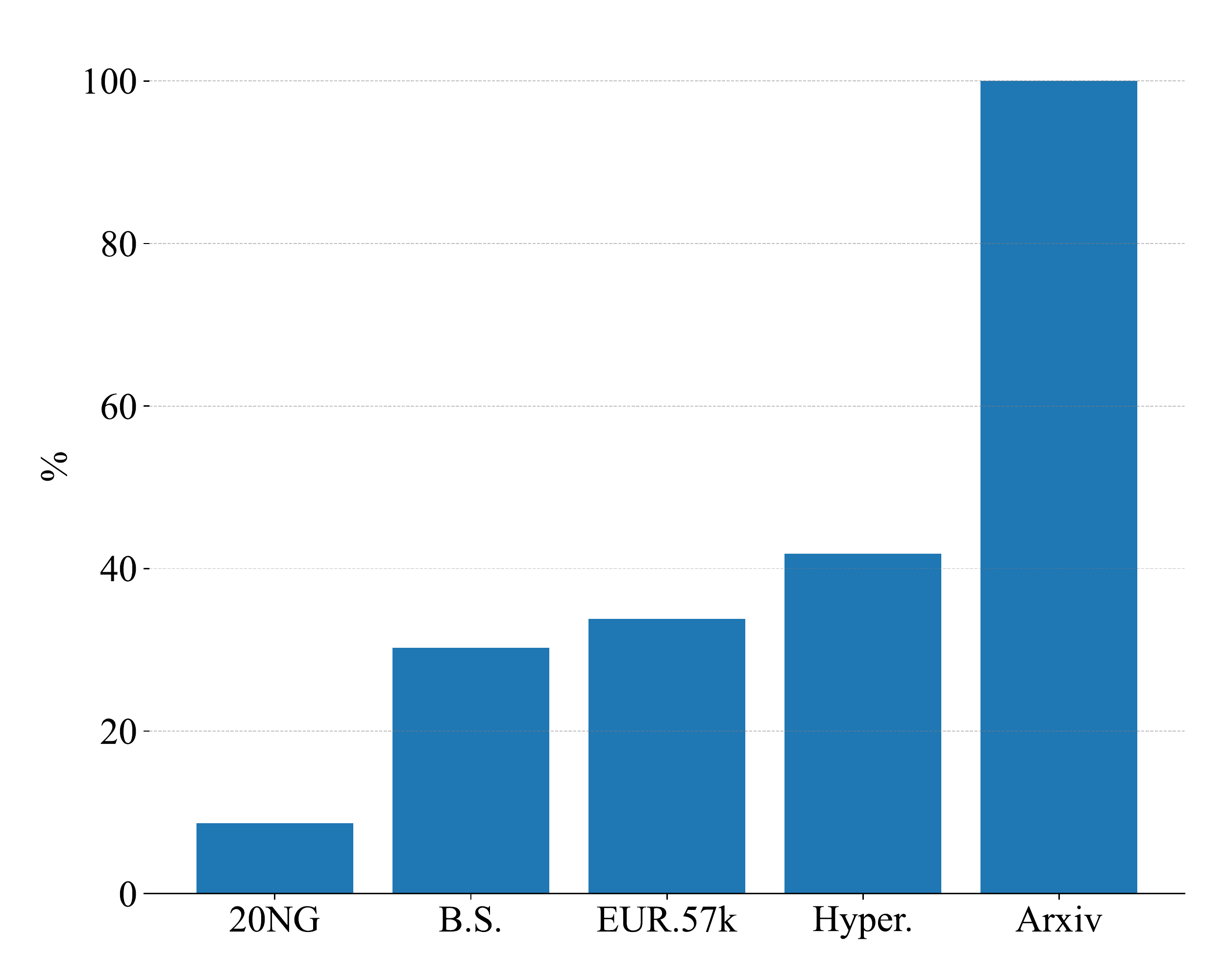}}
    \caption{Statistics of long document benchmark datasets. In (a), the right y-axis is for the Arxiv dataset, and the left y-axis is for the rest datasets.}
    \label{figure-length-boxplot}
\end{figure*}

For the \emph{TC task}, we attempt to test the model performance on datasets with various document lengths. Specifically, we extend the benchmarks from \citet{DBLP:conf/acl/ParkVS22} by adding the long-text dataset Arxiv and the short-text dataset AGNews. The extended benchmarks include (1) \textbf{AGNews}\footnote{\url{http://groups.di.unipi.it/~gulli/AG\_corpus\_of\_news\_articles}}, (2) \textbf{20NewsGroups} \cite{DBLP:conf/icml/Lang95}, and (3) \textbf{Arxiv} \cite{DBLP:journals/access/HeWLFW19} for multi-class classification; (4) \textbf{Book Summary} \cite{DBLP:conf/acl/ParkVS22,DBLP:journals/corr/abs-1305-1319} (\emph{abbr.} \textbf{B.S.}) and (5) \textbf{EURLEX-57K} \cite{chalkidis-etal-2019-large} (\emph{abbr.} \textbf{EUR.57K}) for multi-label classification; and \textbf{Hyperpartisan} \cite{kiesel-etal-2019-semeval} (\emph{abbr.} \textbf{Hyper.}) for binary classification. Figure \ref{figure-length-boxplot} depicts the text length distribution and the long-text ratio of the benchmark datasets. For a fair comparison, following \citet{DBLP:conf/acl/ParkVS22}, we report micro-F1 for multi-label classification and accuracy for binary and multi-class classification.

For the \emph{LM task}, we adopt the commonly-used dataset \textbf{WikiText-103}\footnote{\url{https://blog.salesforceairesearch.com/the-wikitext-long-term-dependency-language-modeling-dataset}} \cite{DBLP:conf/iclr/MerityX0S17} and report the perplexity score following the baselines.

For the \emph{NER task}, we experiment on two widely-adopted English datasets: \textbf{OntoNotesV5.0}\footnote{\url{https://catalog.ldc.upenn.edu/LDC2013T19}} (\emph{abbr.} \textbf{OntoV5}, average length is 77.5) and \textbf{CoNLL2003} \cite{tjong-kim-sang-de-meulder-2003-introduction} (average length is 63.4). Noted that both datasets consist of short texts with an average length shorter than 100, as there are no available NER datasets of long documents. Accordingly, we adopt a small window size for the NER task to test the effectiveness of the recurrent architecture. We use \textit{conlleval}\footnote{\url{https://www.clips.uantwerpen.be/conll2002/ner/bin/conlleval.txt}} to measure the model performance and report the F1 score following the baselines.

\begin{table*}[ht]
\centering
\begin{threeparttable}
\begin{tabular}{lccccccc}
\toprule

        \multirow{2}{*}{\textbf{Model}}   & \multicolumn{1}{c}{AGNews} & \multicolumn{1}{c}{20NG} & 
        \multicolumn{1}{c}{B.S.} & \multicolumn{1}{c}{Hyper.} & \multicolumn{1}{c}{EUR.57K} & \multicolumn{1}{c}{Arxiv} & \multirow{2}{*}{Avg.}\\

        \cmidrule{2-7}
        &  
          \multicolumn{1}{c}{Acc.} & \multicolumn{1}{c}{Acc.} & 
        \multicolumn{1}{c}{F1(micro)} & \multicolumn{1}{c}{Acc.} &
        \multicolumn{1}{c}{F1(micro)} &  \multicolumn{1}{c}{Acc.} \\


\midrule
BiLSTM+GloVe & $93.34$ & $77.97$ & $49.90$  & $90.77$  & $65.00$  & $81.28$   & $76.38$ \\ 
\midrule
BERT   & $93.80$   & $84.79 ^\dagger$  &  $58.18 ^\dagger$  & $92.00 ^\dagger$   & $73.09 ^\dagger$  & $82.00$     &  $80.64$   \\

Longformer   & $93.22$  & $83.39 ^\dagger$  & $56.53 ^\dagger$      & $95.69 ^\dagger$   &  $54.53 ^\dagger$  &  $84.24$    & $77.93$  \\

ToBERT      & $93.80$   & $\mathbf{85.52} ^\dagger$ & $58.16 ^\dagger$    &  $89.54 ^\dagger$  &  $67.57 ^\dagger$   &  $83.75$   & $79.72$     \\

CogLTX   & $93.68$    & $84.63 ^\dagger$  & $58.27 ^\dagger$ & $94.77 ^\dagger$  &  $70.13 ^\dagger$  &  $83.56$   & $80.84$   \\

\midrule
RAN+Random   & $91.70$ & $78.88$  & $50.52$  & $93.85$ & $66.59$    & $80.08$  & $76.94$ \\
RAN+GloVe   & $93.46$ & $79.16$  & $51.58$  & $95.38$ & $67.21$  & $83.36$  & $78.36$ \\
RAN+Pretrain    & $\mathbf{93.83}$ & $85.41$  & $\mathbf{58.43}$ & $\mathbf{96.92}$  & $\mathbf{73.94}$  & $\mathbf{85.92}$  & $\mathbf{82.41}$            
\\ \bottomrule
\end{tabular}
\end{threeparttable}
\caption{Results on long document benchmarks for the classification task. $\dagger$ indicates results retrieved from \citet{DBLP:conf/acl/ParkVS22}. The rest results are from our implementation based on the official code. For the pre-trained word embedding GloVe, we use the embedding model glove.6B.300d\footnote{https://nlp.stanford.edu/data/glove.6B.zip}. \emph{Acc.} is accuracy score. \emph{Avg.} stands for the average numerical results. All the reported results are in percentage (\%). }
\label{table-long-clf-results}
\end{table*}

\subsection{Implementation Details}
The primary experiments in Section \ref{sec-main-result} were conducted using the NVIDIA A100 GPU, while the remaining experiments were conducted using the NVIDIA Titan X GPU (12G memory). The code was implemented using TensorFlow and Keras. By default, we used two layers of RAN, with a window size of 256 for the TC and LM tasks and 64 for the NER task. The head number of pMHSA is set to 12, and the head size is 768. We trained the models for different tasks using the Adam optimizer \cite{Kingma2015Adam} by optimizing the corresponding objective function. For pretrained and non-pretrained RAN models, we set the learning rate to $2e-5$ and $3e-4$, respectively.

\subsection{Main Results}
\label{sec-main-result}
\subsubsection{Long Text Classification}

We compare RAN with baselines on the long-text classification task, including \textbf{BiLSTM} \cite{hochreiter1997long-lstm} and pretrained language models such as \textbf{BERT} \cite{DevlinCLT19BERT}, \textbf{Longformer} \cite{longformer-iz-2020}, \textbf{ToBERT} \cite{tobert-raghavendra-2019}, and \textbf{CogLTX} \cite{cogltx-ding-2020}.
For a comprehensive review, we adopt different initialization methods for RAN parameters. \textbf{RAN+Random} indicates the weights of RAN are randomly initialized. \textbf{RAN+GloVe} stands for using the GloVe embedding \cite{pennington2014glove} as word representation. \textbf{RAN+Pretrain} is the RAN pretrained on the MLM task, following settings in \citet{DevlinCLT19BERT,roberta-liu-2019}. We pretrained RAN on the BookCorpus \cite{Zhu_2015_ICCV} (5GB) and C4 \cite{DBLP:journals/jmlr/RaffelSRLNMZLL20} (RealNews-like subset, 15GB). 

We present the results of long document benchmarks in Table \ref{table-long-clf-results}. In general, the pretrained RAN achieves the five best results among the six benchmarks except for the 20NG dataset and outperforms all the baselines regarding the average score. Note that the pretrained RAN has only $96$M parameters which are fewer than other pretrained baselines, suggesting that RAN is more efficient and scalable than the baselines. Particularly, the pretrained RAN achieves a $2.2\%$ improvement compared with ToBERT on the super-long text dataset Arxiv, demonstrating the superiority of RAN in handling long documents. 

It is worth noticing that the average performance of RAN is higher than that of the chunking-based ToBERT and the document summarization model CogLTX. These two models drop essential information in the chunking and summarizing processes, while RAN can preserve the sequence information with the help of the well-designed recurrent and memory review mechanisms. Moreover, the pretrained RAN achieves the best result on the short-text dataset AGNews, indicating that RAN also performs well in short-text tasks. 

Remarkably, even without pretraining, RAN can still yield competitive performance. For example, the randomly initialized RAN achieved better results than BiLSTM with pretrained GloVe word embedding. RAN with GloVe embedding outperforms pretrained BERT and ToBERT on the accuracy of the Hyper dataset and Longformer on average score. Such observations illustrate that RAN is effective for text encoding and flexible in adopting different initialization methods for various scenarios. It also suggests that the recurrent attention-based architecture of RAN is more powerful than the recurrent architecture of LSTM in modeling texts.

\subsubsection{Language Modeling}
The self-attention-based RAN can be employed for LM. Extensive experiments are conducted to evaluate RAN on language modeling. To avoid information exposure, we apply the causal attention mask to ensure the prediction for $i$-th position only depends on the known outputs before $i$, following \citet{NIPS2017_transformer}. We compare RAN with widely-adopted baselines that are shown in Table \ref{table-language-modeling-results}. The compared models have the same vocabulary and parameter sizes, and the parameters are randomly initialized. The experiment settings follow \citet{zhong2022training}. Observing the results, we notice that RAN achieves the state-of-the-art result on the WikiText-103 dataset with $22.76$ perplexity. It suggests that RAN is efficient in handling the sequence generation task.

\begin{table}[ht]
\small
\centering
\begin{threeparttable}
\begin{tabular}{lcc}
\toprule
Model & \multicolumn{1}{c}{\#Params} & \multicolumn{1}{c}{PPL$\downarrow$} \\ 

\midrule
LSTM \cite{DBLP:conf/iclr/GraveJU17}           & 150M       & $48.70$     \\
TransformerXL   \cite{DBLP:conf/acl/DaiYYCLS19}         & 150M       & $24.00$  \\
B.R. Trans. \cite{blockformer-delesley-2022}  & 150M &\ \ $39.48 ^\dagger$  \\
Transformer \cite{zhong2022training}  & 150M       & $29.14$      \\
Com. Trans.\cite{zhong2022training}  & 150M       & $24.56$      \\
$\infty$-former \cite{zhong2022training} & 150M       & $24.22$      \\
TRIMELM    \cite{zhong2022training}     & 150M       & $25.60$     \\
\midrule
RAN             & 150M     & $\mathbf{22.76}$  \\ 
\bottomrule
\end{tabular}
\end{threeparttable}
\caption{Results of the LM task on the WikiText-103 dataset. Note that the parameter size for language modeling is much larger than that for classification tasks (96M) as we used the same vocabulary for all baselines for a fair comparison. $\downarrow$ means the result is the lower the better. $\dagger$ denotes that the result is from our implementation of the official code. }
\label{table-language-modeling-results}
\end{table}


\subsubsection{Named Entity Recognition}
The NER task is a common information extraction task, and we conduct experiments on the NER task to test RAN for information extraction. As the available NER datasets contain mostly short texts, we set the window size to 64 to test the effectiveness of RAN's recurrent structure. We compare with the following widely-used baselines: \textbf{ID-CNN} \cite{DBLP:conf/emnlp/StrubellVBM17}, \textbf{LSTM} \cite{DBLP:conf/coling/GhaddarL18}, \textbf{LSTM-CNN} \cite{DBLP:conf/aaai/LiFM20}, \textbf{ELMo} \cite{DBLP:conf/naacl/PetersNIGCLZ18}, and \textbf{BERT} \cite{DevlinCLT19BERT}. As shown in Table \ref{table-ner-results}, we notice that RAN consistently outperforms LSTM-based baselines. Specifically, RAN without pretraining achieves $0.5\%$ and $0.3\%$ improvement compared with BERT on both datasets, indicating that the well-designed GPC vector is effective in handling information extraction of long sequences. Both the NER and LM tasks are sequential tasks, and the results demonstrate that RAN is effective in sequence modeling.

\begin{table}[htpb]
\small
\centering
\begin{threeparttable}
\begin{tabular}{lcc}
\toprule
Model & \multicolumn{1}{c}{OntoV5} & \multicolumn{1}{c}{CoNLL2003} \\ 

\midrule
ID-CNN \cite{DBLP:conf/emnlp/StrubellVBM17}          & $86.84$       & $90.54$     \\

LSTM \cite{DBLP:conf/coling/GhaddarL18}         & $87.95$       & $91.73$     \\

LSTM-CNN \cite{DBLP:conf/aaai/LiFM20} & $88.40$       & $-$         \\

ELMo \cite{DBLP:conf/naacl/PetersNIGCLZ18} & $-$ & $92.22$ \\

BERT \cite{DevlinCLT19BERT}           & \ $88.88 ^\dagger$       & $92.40$     \\

\midrule
RAN  ($W=64$)           & $\mathbf{89.38}$       & $\mathbf{92.68}$  \\ 
\bottomrule
\end{tabular}
\end{threeparttable}
\caption{F1 score of the NER task. The results of the baselines are retrieved from the original paper. $\dagger$ denotes results from our implementation by the official code.}
\label{table-ner-results}
\end{table}

\subsection{Ablation Study}
We conducted an ablation study to investigate the significance of each component of our proposed model on the Arxiv dataset. The results are presented in Table \ref{table-ablation-results}. 

In the first ablation model, we substituted the max pooling layer depicted in Eq. \ref{eq:mxpool} with an average pooling layer, resulting in a $1.12\%$ drop in Accuracy. Moreover, our findings show that the residual connection between two windows is essential to alleviate gradient vanishing. When we removed it from RAN, the performance drops approximately $1.6\%$. We also ablated the rotary positional encoding from RAN, which leads to a $1.23\%$ performance drop. 

When the memory review mechanism of RAN was removed in the last ablation model, the result shows the most significant drop compared with other ablation models. RAN without the memory review mechanism suffers a $2.5\%$ performance drop. Such an observation indicates that mitigating information forgetting in processing long documents is crucial, and our proposed memory review mechanism is effective in preserving long-distance memory over the sequence. 

In general, the ablation study demonstrates the significance of each component in our proposed RAN model and highlights the importance of the memory review mechanism in processing long documents. Particularly, our observation accentuates the importance of maintaining long-distance memory in long document processing.

\begin{table}[ht]
\small
\centering
\begin{threeparttable}
\begin{tabular}{lccc}
\toprule
Model & \multicolumn{1}{c}{Accuracy (\%)}  &  $\Delta$ (\%)\\ 

\midrule
RAN+GloVe             & $\mathbf{83.36}$  & \\ 
\midrule
w/ avg pool & $82.24$ & $-1.12$ \\

w/o residual connection  & $81.76$  & $-1.60$ \\

w/o memory review      & $80.85$  & $-2.51$ \\

w/o rotary position     & $82.13$  & $-1.23$ \\
\bottomrule
\end{tabular}
\end{threeparttable}
\caption{Results of ablation models on the Arxiv dataset.}
\label{table-ablation-results}
\end{table}




\subsection{Discussion}
\subsubsection{Scalability Analysis of RAN}
\begin{figure}[t]
    \centering
    \includegraphics[width=\textwidth]{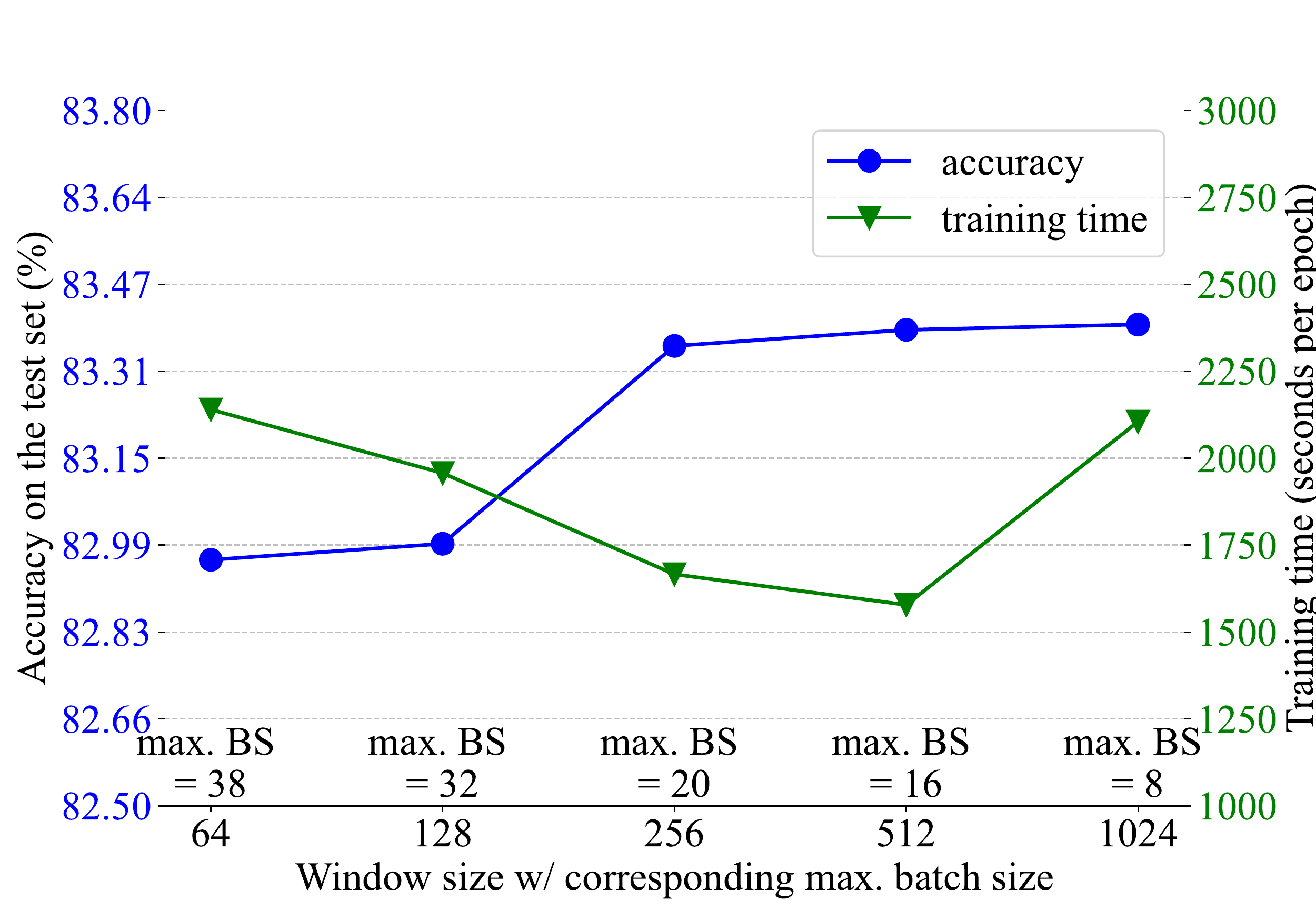}
    \caption{The accuracy of the test set (the blue line with $\bullet$) and the training time (in seconds per epoch, the green line with $\mathsmaller{\blacktriangledown}$) of RAN + Glove with different RAN window sizes on the Arxiv dataset. For each window, the results are obtained with the maximum batch size (max. BS) allowed on a Titan X GPU.}
    \label{different-windows-result-figure}
\end{figure}

This section discusses the scalability of RAN in actual implementation. The window size $W$ determines the number of tokens that are encoded by the attention block. In theory, RAN with a larger window size can yield better performance, as we have fewer windows and less information loss when iterating over the windows. However, given the quadratic memory complexity, the hardware capacity limits the maximum batch size that can be used in training and hence imposes a ceiling to the performance improvement. We have conducted additional experiments with RAN+Glove on the Arxiv dataset. Figure \ref{different-windows-result-figure} depicts the accuracy of the test set and the training time per epoch of RAN with different window sizes. Results of each configuration are obtained with the maximum batch size runnable on the GPU. As expected, when window size increases, the accuracy also gains continuous improvements, albeit minor. The accuracy curve begins to flatten out when the size exceeds 256, partially due to the decreasing maximum batch size. Such an observation indicates the performance is approaching the bottleneck caused by the hardware capacities. 

On the other hand, the V-shape curve observed in the training time is an intriguing sign, and we attribute it to the different time complexities of recurrent and self-attention operations of RAN. Although computing self-attention is of quadratic time, it is significantly accelerated owing to the tensor computation on GPU. In contrast, the recurrent component involves tensor manipulations, such as splitting and concatenation, and thus takes more time. Therefore, a smaller window size will lead to a longer training time as more recurrent operations need to perform. When the window size is large enough, the quadratic time of self-attention becomes significant and dominates the overall spent time. Hence, the green curve bounces back when the window size is 1024. Moreover, when the window size is even larger, such as 2048, the model becomes too large to be loaded on the GPU, and training becomes infeasible. 

Furthermore, we compare the training time of pretrained RAN with other pretrained and non-pretrained baselines on the Arxiv dataset. The results in Table \ref{table-different-trainingtime-results} indicate that the proposed RAN is highly scalable and efficient. Notably, RAN has six times the parameter size of LSTM but has a shorter training time, which indicates our devised recurrent attention structure is more efficient than the LSTM-style gate mechanism.

\begin{table}[htbp]
\small
\centering
\begin{threeparttable}
\begin{tabular}{lrr}
\toprule
Models & \multicolumn{1}{c}{\#Params} & \multicolumn{1}{c}{Time (s/epoch)}  \\ 

\midrule
LSTM (w/o pretrain)    & 15M       & $7,947$     \\

LongFormer (w/ pretrain)  & 148M       & $26,536$   \\

ToBERT  (w/ pretrain)  & 110M       & $6,568$       \\

CogLTX  (w/ pretrain) & 110M & $21,032$ \\
RAN (w/ pretrain) & 96M & $5,393$\\

\bottomrule
\end{tabular}
\end{threeparttable}
\caption{Training time comparison (in seconds per epoch) on Titan X GPU of different models on Arxiv dataset.}
\label{table-different-trainingtime-results}
\end{table}


\subsubsection{The number of RAN Layers}

Similar to RNNs, RAN layers can be stacked to build a deep architecture. We adopt a serial manner to pass the previous layer's GPC output as the input to the subsequent hidden RAN layers. The GPC output at the last RAN layer will be concatenated with the following window input. Intuitively, with more RAN layers, the model will contain more parameters and is promising to produce higher performance. We compare the RANs with different depths and list the results in Table \ref{table-different-layers-results}. As expected, the accuracy improves as the number of layers increases. However, the average training time will significantly increase due to the serial connection between layers, and the improvements become marginal. Therefore, to balance the performance and the time consumption, we adopt the two-layer RAN in this paper by default. This also implies that the results presented in this paper could be further enhanced by increasing the depth of the RAN.

\begin{table}[ht]
\small
\centering
\begin{threeparttable}
\begin{tabular}{ccrc}
\toprule
Layers & \multicolumn{1}{c}{\#Params} & \multicolumn{1}{c}{Time (s/epoch)} & \multicolumn{1}{c}{Accuracy (\%)} \\ 

\midrule
1    & 15M       & $678$ & $83.14$     \\

2   & 17M       & $1,311$ & $83.36$    \\

3  & 19M       & $2,186$  & $83.76$        \\

4 & 21M & $2,967$ & $83.93$\\

\bottomrule
\end{tabular}
\end{threeparttable}
\caption{Results of RAN+GloVe with different layers on Arxiv dataset. \emph{Time} is the average training time (in seconds per epoch).}
\label{table-different-layers-results}
\end{table}


\section{Conclusion \& Future Work}

This paper has presented a novel RAN architecture for long-text modeling that combines the advantages of both recurrent and self-attention networks. The use of a positional multi-head attention mechanism and GPC vector enhances the model's performance by capturing both local and global dependencies in the input sequence. Our ablation study also highlights the critical role of residual connection and memory review mechanisms in preserving long-distance memory. 

With the well-designed recurrent self-attention mechanism, RAN's training can be accelerated by parallel computing on a GPU, making it highly efficient and scalable. We have conducted extensive experiments on TC, NER, and LM tasks. The extensive experiments demonstrate the effectiveness of the proposed RAN model on both classification and sequential tasks.

The flexibility and scalability of our proposed RAN make it a promising choice for future research, with broad potential applications in translation, summarization, conversation generation, and large language models. Additionally, we plan to extend the RAN to tasks involving multi-modality input and output like audio and video, to exploit RAN's long sequence handling capacity in different fields.


\section{Limitations}
The proposed model, Recurrent Attention Network (RAN), effectively models long sequential data by propagating information window-by-window through the sequence via its well-designed recurrent architecture. However, the multi-head self-attention applied to each window is still limited to local attention, which prevents it from providing a global dependency relationship for the entire sequence. This limitation restricts RAN's application in scenarios where a global dependency relationship is necessary, such as visualizing attention weights for the entire document via a heatmap. This limitation potentially reduces the interpretability of the model, although it does not affect the model's performance. Hence, exploring ways to incorporate global attention mechanisms into the RAN architecture is a promising research direction to improve its interpretability and expand its range of applications.



\section*{Acknowledgments}
Xianming Li, Xiaotian Luo, Xing Lee, and Yingbin Zhao's work has been supported by Ant Group. Zongxi Li's work has been supported by a grant from Hong Kong Metropolitan University (Project Reference No. CP/2022/02). Haoran Xie's work has been supported by the Direct Grant (DR23B2) and the Faculty Research Grant (DB23A3) of Lingnan University, Hong Kong. Qing Li's work has been supported by the Hong Kong Research Grants Council through the Collaborative Research Fund (Project No. C1031-18G). We thank the anonymous reviewers for their careful reading of our manuscript. Their insightful comments and suggestions helped us improve the quality of our manuscript.

\bibliography{anthology,custom}
\bibliographystyle{acl_natbib}

\appendix



\end{document}